\documentclass[11pt]{article}
\usepackage{acl2016}
\usepackage{times}
\usepackage{latexsym}
\usepackage{tikz}

\usetikzlibrary{positioning,arrows,decorations.pathreplacing,shapes}

\usepackage{amssymb}
\usepackage{mathtools}
\usepackage{amsmath}
\usepackage{float}

\usepackage{times}
\usepackage{enumerate}

\aclfinalcopy 

\title{Self-organization of vocabularies under different interaction orders}

\author{Javier Vera\\
	    Facultad de Ingenier\'ia y Ciencias\\
	    Universidad Adolfo Ib\'a\~{n}ez\\
	    Avda. Diagonal Las Torres 2640, Pe\~{n}alol\'en, Santiago, Chile\\
	    {\tt jxvera@gmail.com}
	 }

\date{}

\begin{document}

\maketitle

\begin{abstract}
Traditionally, the formation of vocabularies has been studied by agent-based models (specially, the Naming Game) in which random pairs of agents negotiate word-meaning associations at each discrete time step. This paper proposes a first approximation to a novel question: To what extent the negotiation of word-meaning associations is influenced by the order in which the individuals interact? Automata Networks provide the adequate mathematical framework to explore this question. Computer simulations suggest that on two-dimensional lattices the typical features of the formation of word-meaning associations are recovered under random schemes that update small fractions of the population at the same time. 
\end{abstract}

\section{Introduction}

To what extent on a population of language users the formation of a word-meaning association (the simplest version of a \textit{vocabulary}) is influenced by the order in which the individuals interact? Traditionally, this problem has been studied within agent-based models, specially, the Naming Game \cite{Steels95,Steels96,baronchelli_naming_jstat}, in which random pairs of agents (one speaker, one hearer) negotiate word-meaning associations at each discrete time step. Naturally, in a more realistic scenario at some time frame (for instance, at the same minute) multiple unrelated communicative interactions occur. 

Automata Networks (AN) \cite{Neumann:66,wolfram02} provide the adequate mathematical framework to stress the self-organized nature of the formation of vocabularies \cite{steels2011REVIEW,1742-5468-2011-04-P04006}. AN are extremely simple models where each vertex of a set of vertices (the network) evolves following a local rule based on the states of ``nearby" vertices. At each time step the entire set or a fraction of the set of vertices (even one vertex) is updated. This model is therefore the natural account to describe the influence of the order in which the individuals interact on the consensus of the entire population. 

The main purpose of the definition of an AN model is twofold: (1) to reproduce the typical dynamics of the formation of linguistic conventions (in particular, of the Naming Game); and (2) to describe by computer simulations the dynamics of consensus formation under different orders of communicative interactions. 

The work is organized as follows. Section 2 explains basic notions and the local rules of the AN model. This is followed by computer simulations focused on the dynamics under four interaction orders. Finally, a brief discussion about the relations between the results and the formation of language is presented. 

\section{Automata Networks}

The individuals are located on the vertices of a connected, simple and undirected graph $\mathcal{G}=(P,I)$, where $P=\{1,...,n\}$ is the set of vertices (the population) and $I=\{1,...,m\}$ is the set of edges. In order to constraint the communicative interactions the \textit{neighborhood} of the vertex $u \in P$ is defined as the set $V_u=\{v \in P: (u,v) \in I\}$. The vertex $u$ uniquely interacts with its \textit{neighbors}, located on $V_u$. 

A set of $p$ words $W$ is considered. Each individual is completely characterized by its \textit{state}, which evolves within communicative interactions. The state associated to the individual $u \in P$ is the pair $(M_u,x_u)$, where $M_u \subseteq W$ is the memory to store words of $u$ and $x_u \in M_u$ is a word that $u$ conveys to its neighbors of $V_u$. The set of words conveyed by the neighbors of the vertex $u$ is denoted $W_u=\{x_v: v \in V_u\}$. Since the formation of a language is based only on \textit{local} interactions, the vertex $u$ only accesses to the set $W_u$. In principle it is reasonable to think that the vertex $u$ plays the role of ``hearer" (it hears the words conveyed by the neighbors of $u$), and the vertices of $V_u$ play the role of ``speaker" (they convey words to the vertex $u$). In general communicative interactions in the AN model involve multiple speakers (the neighbors) and one hearer (the central vertex). 

The AN model is the tuple $\mathcal{A}=(\mathcal{G},Q,(f_u:u \in P),\phi)$, where

\begin{itemize}
\item $Q$ is the set of all possible states of the vertices ($Q = \mathcal{P}(W) \times W$, where $\mathcal{P}(W)$ means the set of subsets of $W$);
\item $(f_u:u \in P)$ is the family of \textit{local rules}. The state of a given vertex $u$ evolves taking into account the states of its neighbors of $V_u$; and 
\item $\phi$ is a function, the \textit{updating scheme}, that gives the order in which the vertices are updated. Four updating schemes are considered: the \textit{sequential} scheme, where vertices are updated one by one in a prescribed order (a \textit{permutation} of the set of vertices); the \textit{fully-asynchronous} scheme, where vertices are updated one by one in an uniformly at random order; the \textit{synchronous} scheme, where all vertices are updated at the same time; and the $\alpha$-\textit{asynchronous} scheme, where each vertex is updated with probability $\alpha$ \cite{F}. It is clear that $\alpha=1$ is equivalent to the \textit{synchronous} scheme.  
\end{itemize}

A \textit{configuration} $X(t)$ at $t$ is the set of states $\{(M_u(t),x_u(t))\}_{u \in P}$. The configuration $X(t+1)$ at time step $t+1$ is given by the application of local rules $\{f_u\}_{u \in P}$ to a subset of individuals defined by the updating scheme. There are two special configurations: (1) a \textit{cycle} (a finite periodic set of configurations); (2) and a \textit{fixed point} (a configuration which is invariant under the application of local rules). 

\section{Two actions based on the Naming Game}

To model the formation of consensus the local rule involves two actions which update the state pair $(M_u,x_u)$ of each vertex $u \in P$: 

\begin{itemize}
\item the \textbf{addition (A)} updates $M_u$ by adding words; and 
\item the \textbf{collapse (C)} updates $M_u$ by cancelling all its words, except one of them. 
\end{itemize}

Both actions arise from \textit{lateral inhibition} strategies proposed in the context of the Naming Game \cite{Steels95,steels2011REVIEW}. The \textit{addition} attempts to increase the chance of future successful interactions. The \textit{collapse} is the result of local consensus within interactions (the individuals remove the remaining words in ``successful" conversations). Two simple forms of these actions are considered. First, the individuals add all the unknown words. This means that the vertex $u \in P$ adds any word (conveyed by its neighbors) $x \in W_u$, so that $x \notin M_u$. Second, in the case that $W_u \subseteq M_u$ (all received words belong to the memory), the constraints of the AN model impose the definition of a mechanism that allows to \textit{discriminate} between words. Indeed, the vertex $u$ receives one word from each neighbor of $V_u$. To assign different values to the words each agent is endowed with an internal total order for the set of words (equivalently, if we consider $W\subseteq \mathbb{Z}$ the agents are endowed with the order $<$). The individuals choose to collapse in the minimum word conveyed in the neighborhood, that is, the minimum of the set $W_u$. 

\section{Rules of the model}

\begin{figure}[h!]
\begin{center}
\begin{tikzpicture}
  [scale=.4,auto=left,every node/.style={minimum size=0cm}] 
  [thick,scale=0.6,shorten >=0pt]
 
  \node (n1) at (3,5)  {\large $(\{a,b\},b)$}; 
   \node (n2) at (-0.5,7.5)  {\large $b$};
   \node (n3) at (6.5,7.5)  {\large  $c$};
   \node (n4) at (3,2) {\large $d$};

  \foreach \from/\to in {n1/n2,n1/n3,n1/n4}
    \draw (\from) -- (\to);
    
    
   \node (n5) at (15,5)  {\large $(\{a,b,c,d\},b)$}; 
   \node (n6) at (11.5,7.5)  {};
   \node (n7) at (18.5,7.5)  {};
   \node (n8) at (15,2) {};

  \foreach \from/\to in {n5/n6,n5/n7,n5/n8}
    \draw (\from) -- (\to);
    
    
   \node (n9) at (7,5)  {};
   \node (n10) at (11,5)  {};
   
   \path (n9) edge [->, line width=2pt] node[]{\large \textbf{(A)}} (n10);
   
   
   \node (n11) at (3,-3)  {\large $(\{a,b,c\},b)$}; 
   \node (n12) at (-0.5,-0.5)  {\large $b$};
   \node (n13) at (6.5,-0.5)  {\large $c$};
   \node (n14) at (3,-6) {\large $a$};

  \foreach \from/\to in {n11/n12,n11/n13,n11/n14}
    \draw (\from) -- (\to);
    
    
   \node (n15) at (15,-3)  {\large $(\{a\},a)$}; 
   \node (n16) at (11.5,-0.5)  {};
   \node (n17) at (18.5,-0.5)  {};
   \node (n18) at (15,-6) {};

  \foreach \from/\to in {n15/n16,n15/n17,n15/n18}
    \draw (\from) -- (\to);
    
    
   \node (n19) at (7,-3)  {};
   \node (n20) at (11,-3)  {};
   
   \path (n19) edge [->, line width=2pt] node[]{\large \textbf{(C)}} (n20);

\end{tikzpicture}
\end{center}
\caption{\textbf{Example of the local rule}. All individuals share the order $a<b<c<d$. Suppose that at some time step the vertex $u$ has been choosen (the central vertex). Four individuals participate of the interaction: the central vertex $u$ and its three neighbors of $V_u$. Associated to each action of the local rule two different configurations are showed. For the first row \textbf{(A)}, $W_u=B_u \cup N_u=\{b\} \cup \{c,d\}$. For the second row \textbf{(C)}, $W_u=B_u=\{a,b,c\}$.}
\end{figure}
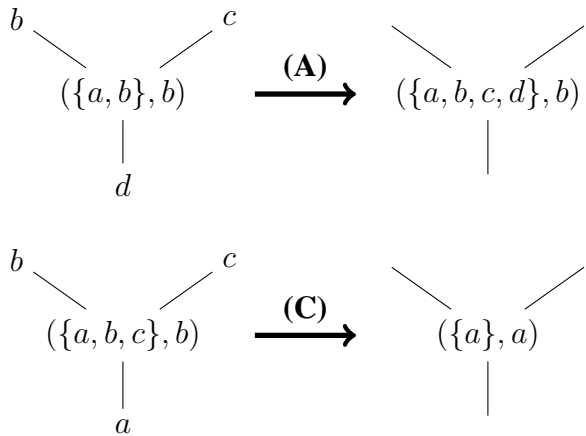

Suppose that at time step $t$ the vertex $u \in P$ has been selected according to one of the updating schemes. $W_u$ is the set of all words conveyed by the neighbors of the vertex $u$. $W_u=B_u \cup N_u$, where $B_u=\{x_v \mid (v \in V_u) \land (x_v \in M_u)\}$ (the set of \textit{known} words) and $N_u=\{x_v \mid (v \in V_u) \land (x_v \notin M_u)\}$ (the set of \textit{unknown} words). The subset $B_u \subseteq W_u$ contains the words conveyed by the neighbors which belong to the memory $M_u$. In contrast, $N_u \subseteq W_u$ contains the words of $W_u$ which do not belong to $M_u$. The local rule $f_u$ reads

\begin{equation*}
 f_u = \left\{ \begin{array}{ll} \textrm{if } \emptyset \neq N_u, &  \textbf{(A) } (M_i \cup N_u,x_u) \\ \textrm{if }  \emptyset = N_u, & \textbf{(C) } (\{\min(B_u)\},\min(B_u))  \end{array} \right.
\end{equation*}


The rule means that in the case that $\emptyset \neq N_u$ (that is, if there are words of $W_u$ that the vertex $u$ does not know) the memory $M_u$ is updated by adding the words of $N_u$. In the other case ($\emptyset = N_u$) the vertex $u$ collapses its memory in the minimum of the set $B_u$ ($B_u=W_u$). Clearly, the conveyed word $x_u$ eventually changes in collapses.

\section{Simulations}

\subsection{Protocol}

To explicitly describe the dynamics of the AN two macroscopic measures are defined \cite{baronchelli_naming_jstat}: the total number of words of the system,

\begin{equation}
n_w(t)=\sum_{u \in P} |M_u|
\end{equation}

where $|M_u|$ is the size of the memory $M_u$; and the number of different words (or synonyms),

\begin{equation}
n_d(t)=|\bigcup_{u \in P} M_u|
\end{equation}

where $\bigcup_{u \in P} M_u$ represents the union of all sets $M_u$, $u \in P$.

The simulation protocol is defined by the following elements. Averages of $n_w(t)$ and $n_d(t)$ over 100 initial conditions where each vertex is associated to a different state of the form $(\{x\},x)$, $x \in W$. Then, $n_w(0)=n_d(0)=n$. Four updating schemes: sequential, fully asynchronous, synchronous and $\alpha$-asynchronous (with $\alpha$ in $\{0.1,0.9\}$). A periodic lattice with Von Neumann neighborhood (four nearest neighbors) with $n=256^2=65536$ vertices for both sequential and fully asynchronous schemes, and $n=64^2=4096$ vertices for both synchronous and $\alpha$-asynchronous schemes.

\subsection{Results}

\begin{figure}[h!]

  \begin{center} 
    
  \includegraphics[scale=0.3]{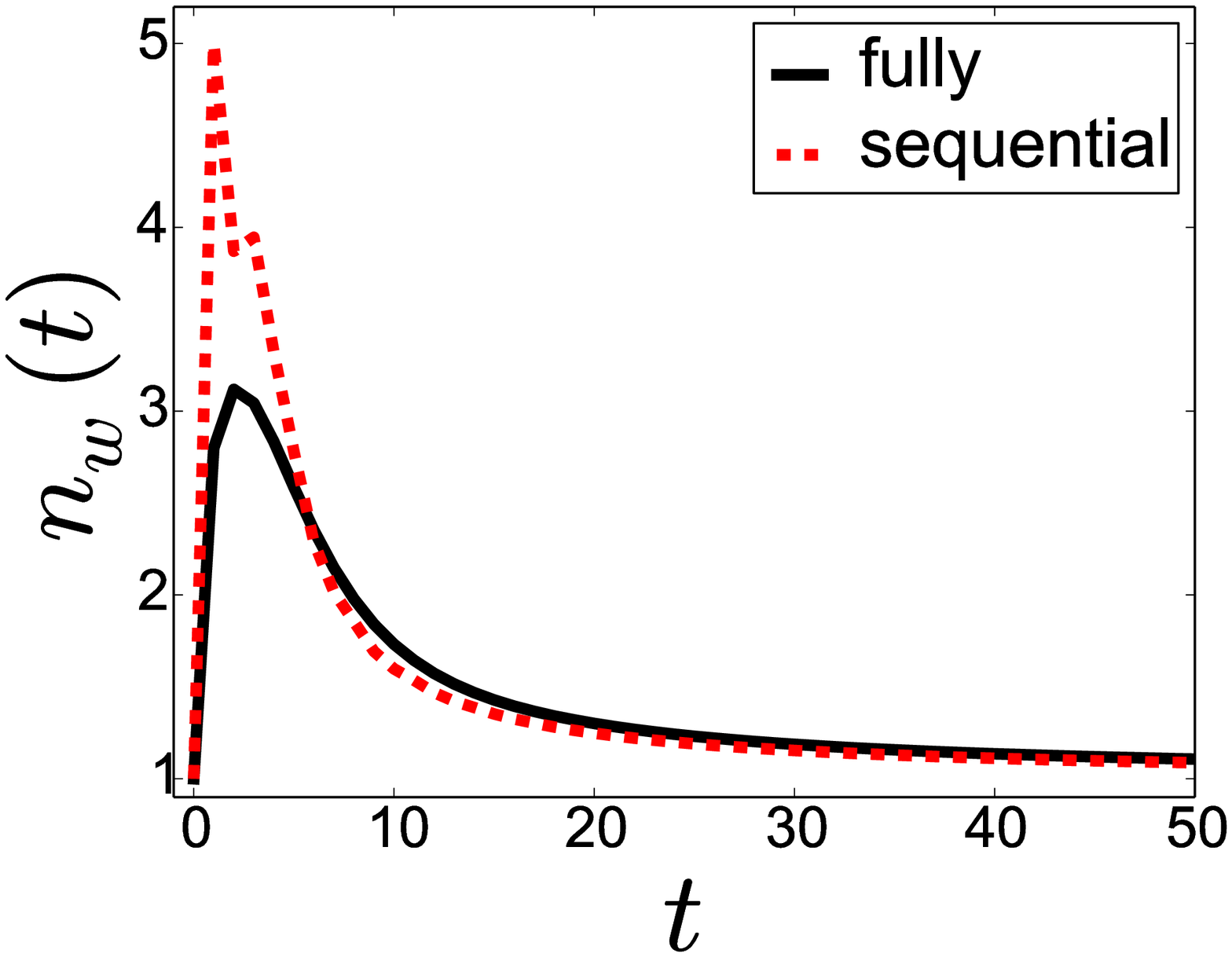}
  \includegraphics[scale=0.3]{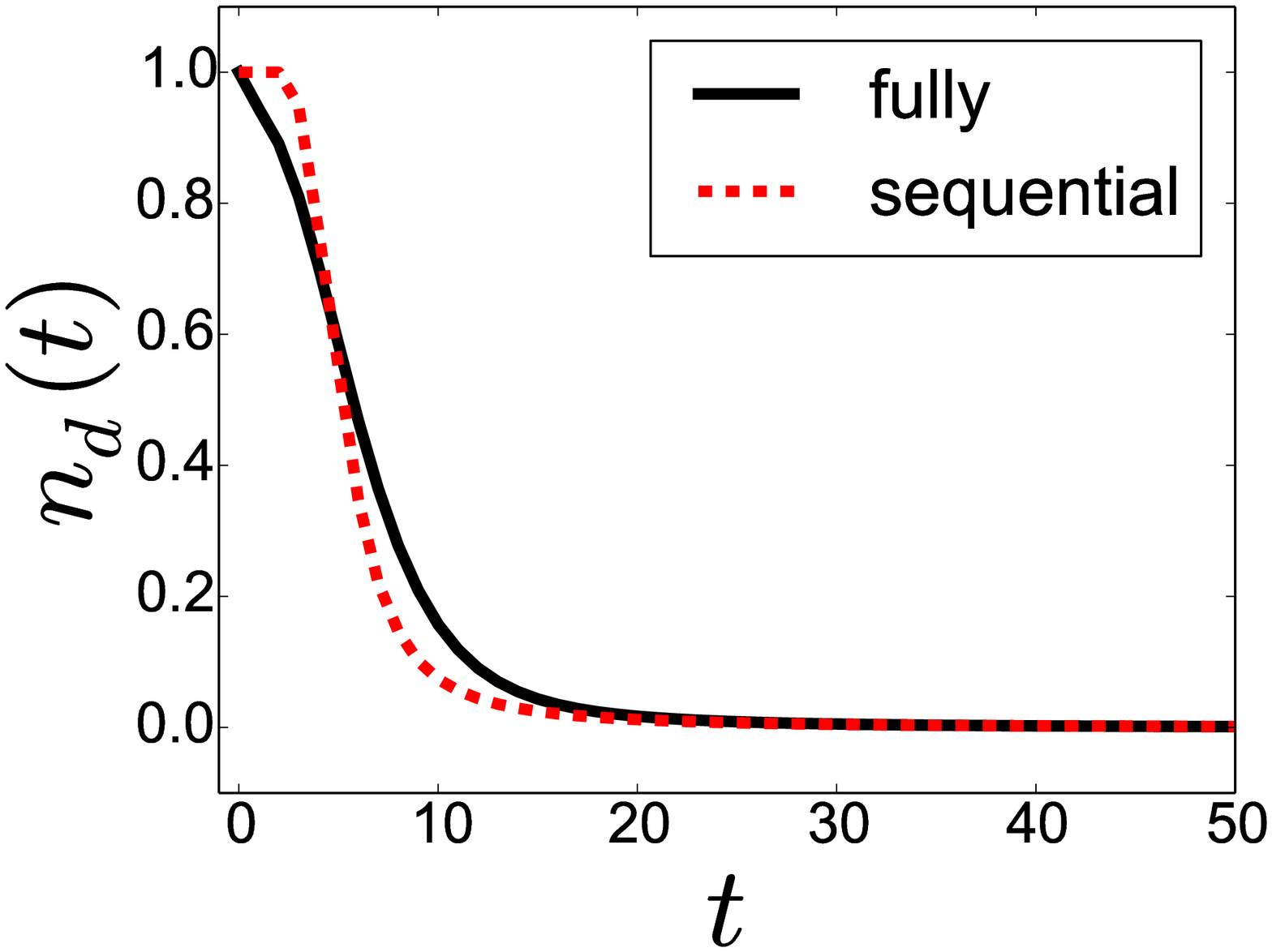}\\

  \end{center}
  \caption{\textbf{Evolution of $n_w(t)$ and $n_d(t)$ under sequential and fully-asynchronous updating schemes}. The population is located on a $n=256^2$ periodic lattice with Von Neumann neighborhood (four nearest neighbors). The simulations run until reach $n_d(t)=1$. \textbf{(top)} $n_w(t)$ versus $t$. \textbf{(bottom)} $n_d(t)$ versus $t$. One step $t$ means $n$ vertex updates. y-axis is normalized by $n$. Only the first $50$ steps are showed. Black lines mean fully asynchronous scheme, whereas red depicted lines mean sequential scheme. Both schemes converge at $t \sim 300$.}

\end{figure}

\begin{figure}[h!]

  \begin{center} 
    
  \includegraphics[scale=0.3]{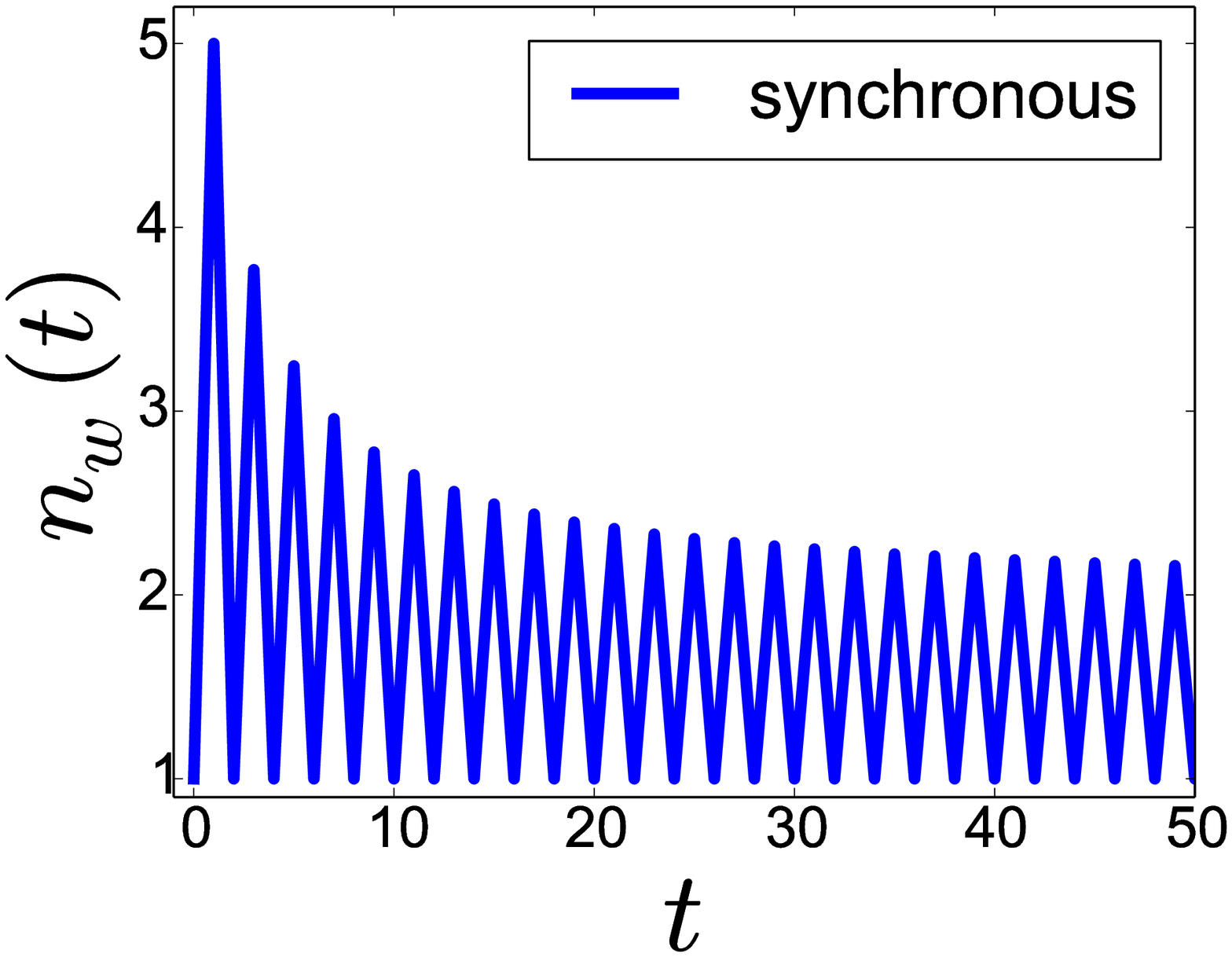}
  \includegraphics[scale=0.3]{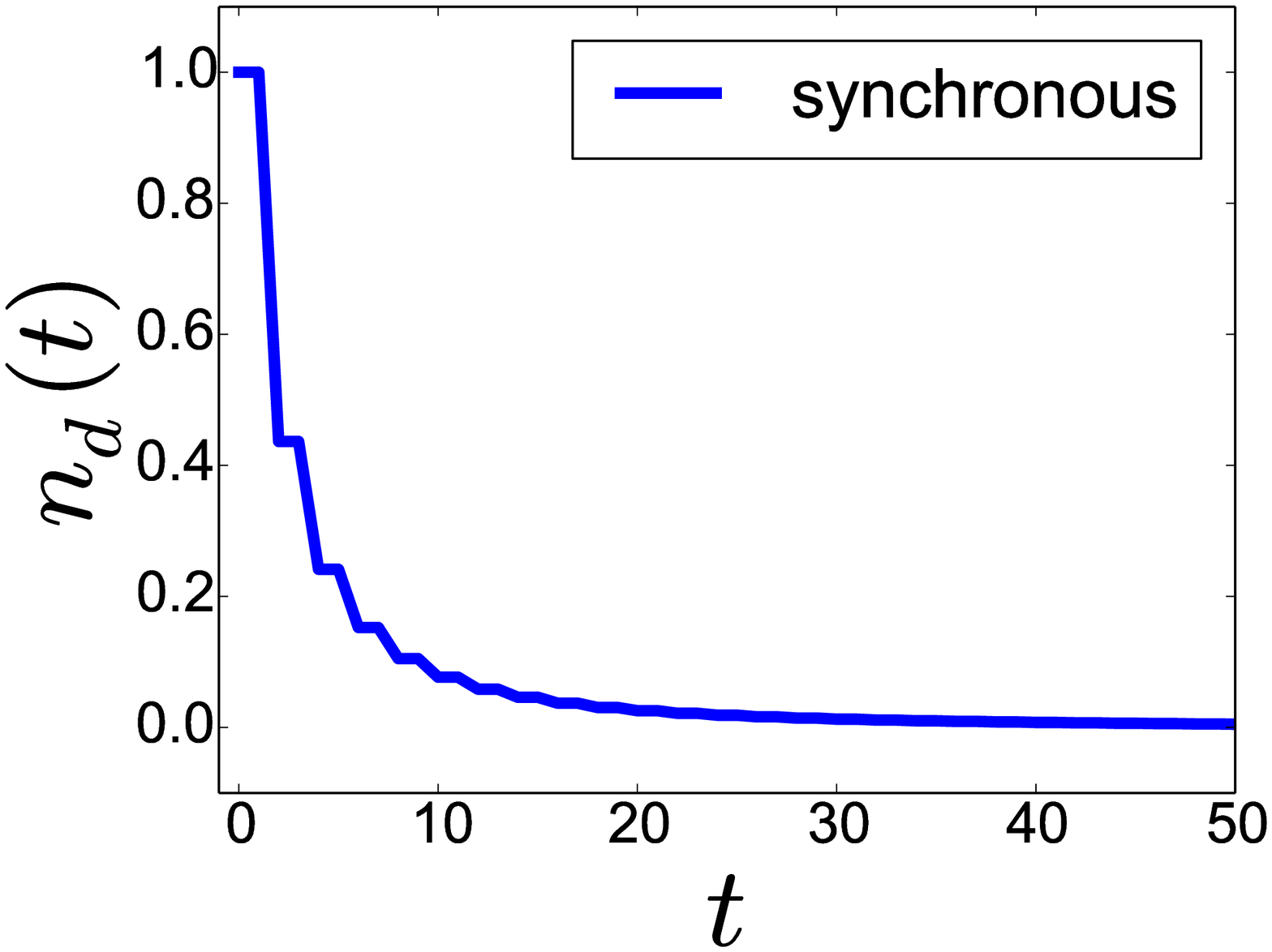}\\
  
  \end{center}
  \caption{\textbf{Evolution of $n_w(t)$ and $n_d(t)$ under synchronous updating scheme}. The population is located on a $n=64^2$ periodic lattice with Von Neumann neighborhood (four nearest neighbors). The simulations run $200$ steps. \textbf{(top)} $n_w(t)$ versus $t$. \textbf{(bottom)} $n_d(t)$ versus $t$. One step $t$ means $n$ vertex updates. y-axis is normalized by $n$. Only the first $50$ steps are showed. The dynamics for all initial conditions leads to cycles of size 2.}

\end{figure}

\begin{figure}[h!]

  \begin{center} 
    
  \includegraphics[scale=0.3]{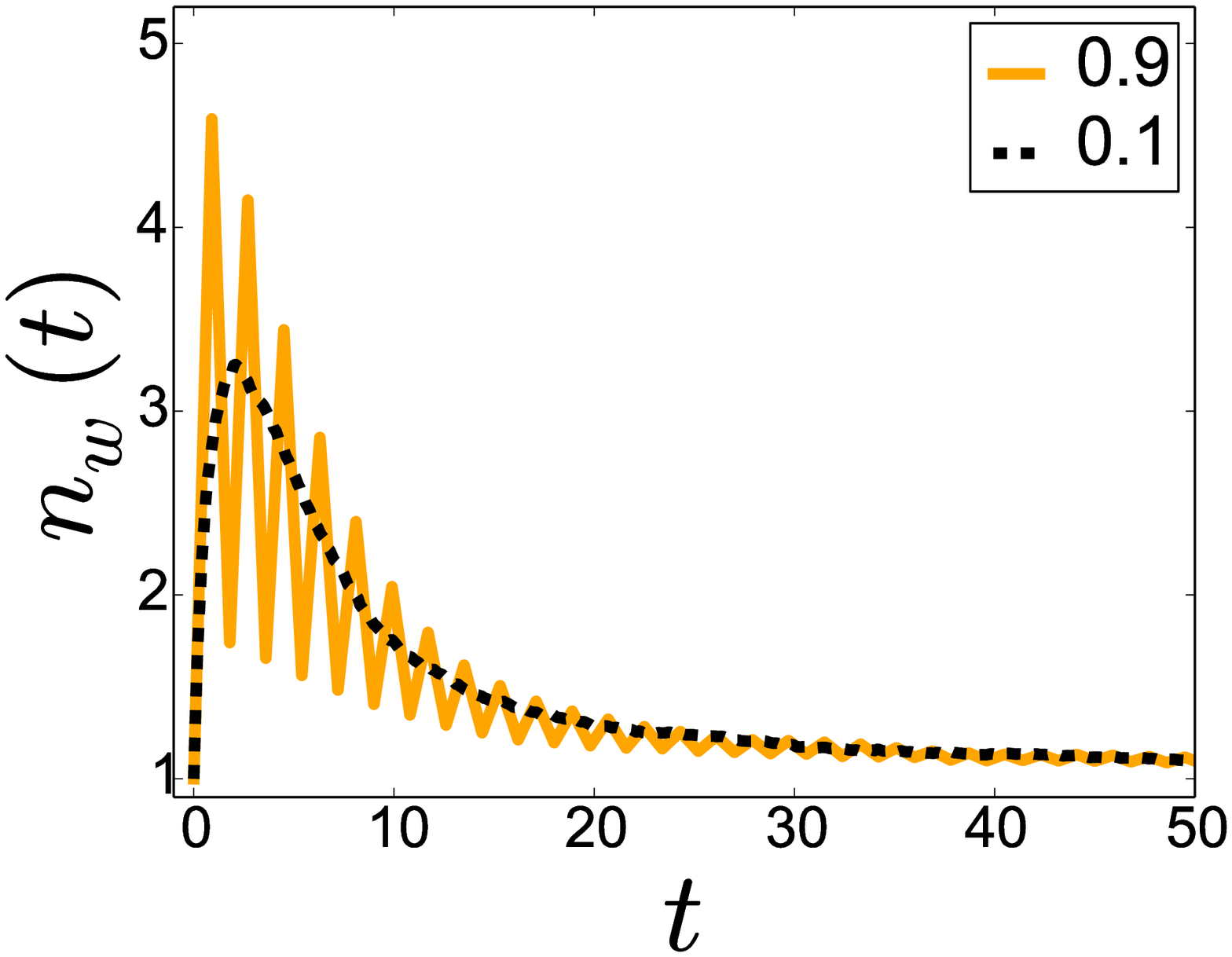}
  \includegraphics[scale=0.3]{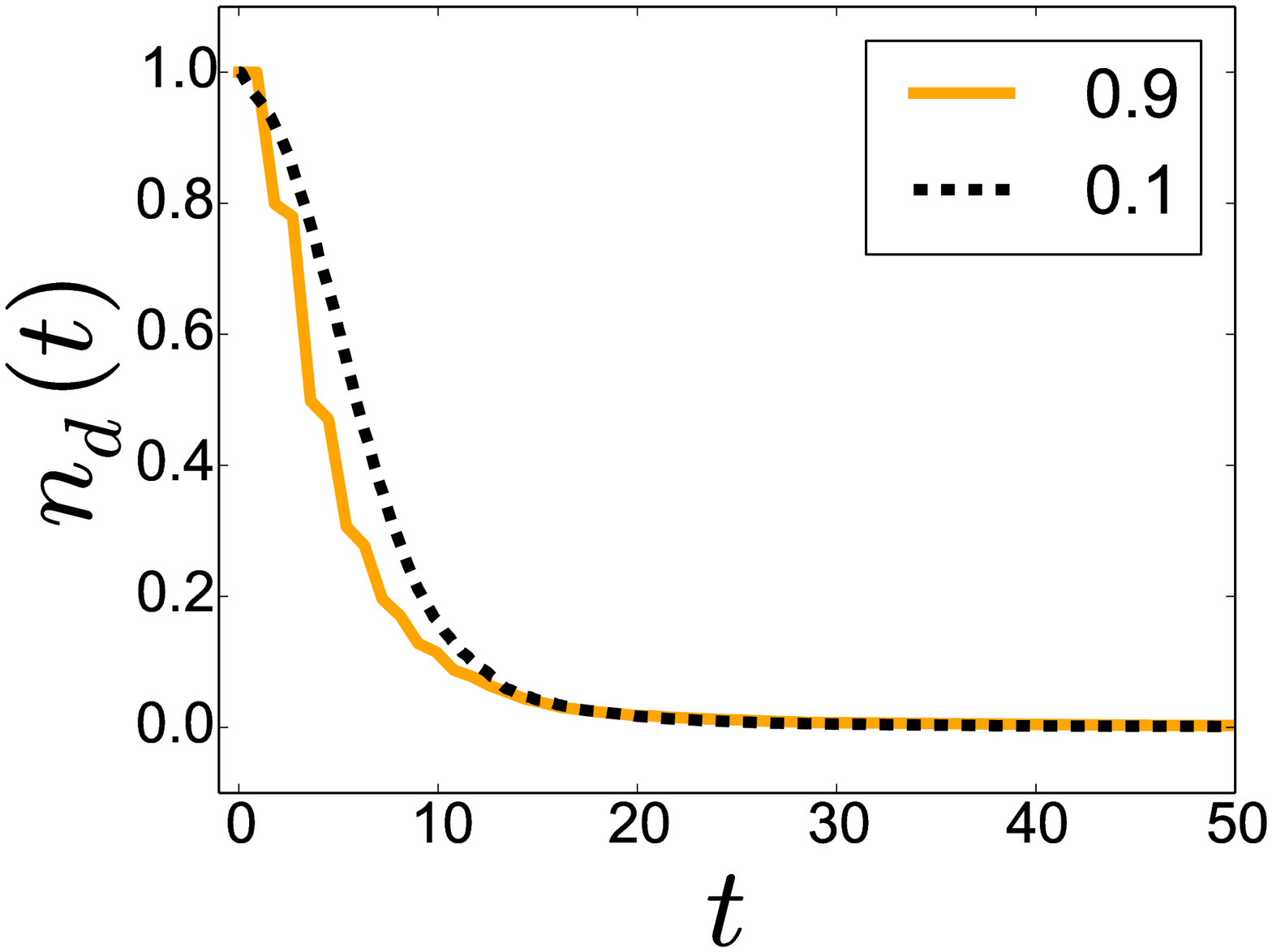}\\
  
  \end{center}
  \caption{\textbf{Evolution of $n_w(t)$ and $n_d(t)$ under $\alpha$-asynchronous updating scheme}. The population is located on a $n=64^2$ periodic lattice with Von Neumann neighborhood (four nearest neighbors). The simulations run $200$ steps. \textbf{(top)} $n_w(t)$ versus $t$. \textbf{(bottom)} $n_d(t)$ versus $t$. $\alpha$ varies from $\{0.1,0.9\}$. x-axis is normalized by $\alpha$ and y-axis is normalized by $n$. Only the first $50$ steps are showed. }

\end{figure}

The dynamics of the AN under fully-asynchronous updating scheme seems to reproduce the typical behavior observed for the Naming Game on low-dimensional lattices \cite{baronchelli_naming_jstat}, as shown in Fig. 2 (black lines). Indeed, the dynamics exhibits three typical domains. First, since at the beginning the individuals convey different words a very fast increasing in $n_w(t)$ and a drastic decreasing in $n_d(t)$ are observed. Then, a peak in the number of words is reached, at $n_w(t) \sim 3n$. Finally, the dynamics enters in a very slow convergence to the consensus configuration, where $n_w(t)=n$ and $n_d(t)=1$. The convergence is reached at $t \sim 300$ steps (only the first 50 steps are showed). 

The dynamics under sequential scheme presents some remarkable aspects, as shown in Fig. 2 (red depicted lines). The evolution of the number of words reaches a very sharp peak of $n_w(t) \sim 5n$. This means that at the peak each individual knows all the words conveyed by its neighbors (notice that each vertex has four neighbors). Another interesting feature of the dynamics is that after the peak the dynamics reaches a local maximum at $t \sim 3$. This fact requires further mathematical explanation. 

Synchronous dynamics is exhibited in Fig. 3 (blue lines). Approximately after $50$ time steps (one step means that all individuals have been updated) the dynamics enters in a periodic behavior with cycles of length 2. Thereby, the number of words $n_w(t)$ oscillates between $n$ and $2n$, whereas $n_d(t)$ converges to 2. In fact, time steps that correspond to even numbers (even steps) imply collapses ($n_w(t)=n$) and odd steps imply additions ($n_w(t)=2n$). Small ``ladder" steps in the decreasing evolution of $n_d(t)$ show that at odd times only additions are allowed (and then the conveyed words remain fixed). 

The dynamics under $\alpha$-asynchronous scheme is exhibited in Fig. 4. The two values of $\alpha$ (0.9 and 0.1) are related to previous observations. First, at $\alpha=0.9$ the dynamics presents oscillations that diminish over time until reach a final consensus fixed point. Second, at $\alpha=0.1$ the dynamics seems to reproduce the behavior observed for the Naming Game on low-dimensional lattices (as the fully-asynchronous scheme). 
 
\section{Conclusion}

This short paper introduces a new theoretical framework to study the development of linguistic conventions, and in general, the formation and evolution of language. Despite that the AN model is an \textit{abstract} approximation to the real problem, the work discusses how on a population of individuals endowed with simple cognitive mechanisms language arises only from local interactions. 

The work proposes two important elements to be discussed: (1) an alternative (mathematical) framework for agent-based studies on language formation; and (2) computer simulations suggesting that on two-dimensional lattices the typical features of the formation of linguistic conventions (as in the Naming Game) are recovered under random schemes that update small fractions of the population at the same time (fully-asynchronous and $\alpha$-asynchronous, associated to $\alpha=0.1$).

\section*{Acknowledgments}
The author likes to thank CONICYT-Chile under the Doctoral scholarship 21140288.

\bibliography{acl2016}
\bibliographystyle{acl2016}

\appendix

\end{document}